\documentclass[graybox]{svmult}

\usepackage{type1cm}        
\usepackage{makeidx}         
\usepackage{graphicx}        
\usepackage{multicol}        
\usepackage{multirow}
\usepackage[bottom]{footmisc}
\usepackage{pdfpages}

\usepackage{newtxtext}       %
\usepackage[varvw]{newtxmath}       
\usepackage{booktabs}
\usepackage{hyperref}

\makeindex             

\begin{document}

\title*{Explainable AI for Classifying UTI Risk Groups Using a Real-World Linked EHR and Pathology Lab Dataset}

\titlerunning{XAI for Classifying UTI Risk Groups Using a Real-World...}
\author{Yujie Dai\orcidID{0009-0006-6082-736X} and\\ Brian Sullivan\orcidID{0000-0002-5275-3931}
and\\ Axel Montout and\\ Amy Dillon and\\ Chris Waller and\\ Peter Acs and\\ Rachel Denholm and\\  Philip Williams and\\ Alastair D Hay and\\ Raul Santos-Rodriguez\orcidID{0000-0001-9576-3905} and\\  Andrew Dowsey\orcidID{0000-0002-7404-9128}}
\authorrunning{Y. Dai et al.}
\institute{Yujie Dai \at University of Bristol, UK \email{yujie.dai@bristol.ac.uk}
\and Brian Sullivan \at University of Bristol, UK \email{brian.sullivan@bristol.ac.uk}
\and Axel Montout \at University of Bristol, UK \email{axel.montout@bristol.ac.uk}
\and Amy Dillon \at University of Bristol, UK \email{amy.dillon@bristol.ac.uk}
\and Chris Waller \at Kettering General Hospital NHS Foundation Trust, UK \email{christopher.waller@nhs.net}
\and Peter Acs \at NHS England, UK \email{peter.acs@nhs.net}
\and Rachel Denholm \at University of Bristol, UK \email{r.denholm@bristol.ac.uk}
\and Philip Williams \at University Hospitals Bristol and Weston NHS Foundation Trust, UK \email{Philip.Williams2@UHBW.nhs.uk}
\and Alastair D Hay \at University of Bristol, UK \email{alastair.hay@bristol.ac.uk}
\and Raul Santos-Rodriguez \at University of Bristol, UK \email{enrsr@bristol.ac.uk}
\and Andrew Dowsey \at University of Bristol, UK \email{andrew.dowsey@bristol.ac.uk}}

\maketitle
\abstract*{The use of machine learning and AI on electronic health records (EHRs) holds substantial potential for clinical insight. However, this approach faces challenges due to data heterogeneity, sparsity, temporal misalignment, and limited labeled outcomes. In this context, we leverage a linked EHR dataset of approximately one million de-identified individuals from Bristol, North Somerset, and South Gloucestershire, UK, to characterize urinary tract infections (UTIs). We implemented a data pre-processing and curation pipeline that transforms the raw EHR data into a structured format suitable for developing predictive models focused on data fairness, accountability and transparency. Given the limited availability and biases of ground truth UTI outcomes, we introduce a UTI risk estimation framework informed by clinical expertise to estimate UTI risk across individual patient timelines. Pairwise XGBoost models are trained using this framework to differentiate UTI risk categories with explainable AI techniques applied to identify key predictors and support interpretability. Our findings reveal differences in clinical and demographic predictors across risk groups. While this study highlights the potential of AI-driven insights to support UTI clinical decision-making, further investigation of patient sub-strata and extensive validation are needed to ensure robustness and applicability in clinical practice. }
\abstract{The use of machine learning and AI on electronic health records (EHRs) holds substantial potential for clinical insight. However, this approach faces challenges due to data heterogeneity, sparsity, temporal misalignment, and limited labeled outcomes. In this context, we leverage a linked EHR dataset of approximately one million de-identified individuals from Bristol, North Somerset, and South Gloucestershire, UK, to characterize urinary tract infections (UTIs). We implemented a data pre-processing and curation pipeline that transforms the raw EHR data into a structured format suitable for developing predictive models focused on data fairness, accountability and transparency. Given the limited availability and biases of ground truth UTI outcomes, we introduce a UTI risk estimation framework informed by clinical expertise to estimate UTI risk across individual patient timelines. Pairwise XGBoost models are trained using this framework to differentiate UTI risk categories with explainable AI techniques applied to identify key predictors and support interpretability. Our findings reveal differences in clinical and demographic predictors across risk groups. While this study highlights the potential of AI-driven insights to support UTI clinical decision-making, further investigation of patient sub-strata and extensive validation are needed to ensure robustness and applicability in clinical practice. }

\runinhead{Keywords} Electronic Health Records, Machine Learning, Explainable AI, Urinary Tract Infections

\section{Introduction}
\label{sec:1}
The integration of electronic health records (EHRs) into health research offers immense potential for generating insights that improve patient care and health outcomes ~\cite{KingStudy,MenachemiReview}. As large-scale health data becomes more accessible, machine learning and AI can reveal patterns and relationships that traditional statistical methods might overlook \cite{RICHTER20181}. However, the use of machine learning and AI with EHR data presents notable challenges, as EHRs are observational datasets that are often heterogeneous, sparse, and temporally misaligned \cite{ghassemi2020review,XIE2022103980}. These challenges are due to variations in clinical outcomes, diverse disease stages or conditions, irregular visit intervals, and inconsistent data recording practices over time that were not designed with machine learning in mind. Moreover, the lack of labeled outcomes, e.g. definitive diagnoses, complicates the development of robust predictive models, particularly in supervised learning. To address these challenges, rigorous data curation and pre-processing methods are essential.

Several publicly available datasets, such as the MIMIC-IV \cite{MIMIC4} and eICU \cite{eICU}, are frequently used in healthcare predictive modeling. While these datasets contain detailed patient information (e.g. demographics, lab results, and clinical notes) they primarily focus on intensive care settings, limiting their applicability to conditions managed predominantly in primary care, such as urinary tract infections (UTIs) \cite{Leckye330}. 

UTIs are one of the most common classes of infectious disease encountered in clinical practice \cite{UTI1}. A number of studies have applied machine learning techniques to predict UTI risk using EHR data. Classification algorithms like logistic regression and decision trees have been employed to predict UTI presence based on structured data fields such as patient demographics and medical history \cite{UTI_pred_C,UTI_Jeng}. Other studies use clustering techniques to segment patients into UTI risk subgroups, revealing patterns that may guide targeted interventions \cite{BARCHITTA202157}. Additionally, \cite{UTI_NLP} incorporated unstructured data via natural language processing to extract symptoms from clinical notes, such as Urinary frequency, to enhance UTI prediction. Reported UTI risk factors include gender (with females at higher risk), advanced age, comorbidities such as diabetes and neurological disorders, and prior history of UTIs which predispose individuals to recurrence \cite{UTI_pred_C,UTI_Jeng,JAKOBSEN2023}.

However, data used in these studies often lacks the breadth of linked primary care, secondary care and laboratory records, potentially limiting the predictive models' depth and accuracy, as many studies rely solely on either primary or secondary care data alone. This can lead to inequities, as risk factors may vary across different population subgroups, with hospitalized patients tend to be more vulnerable. Furthermore, few studies have examined the discrepancies between clinician-based and AI-driven risk assessments for UTIs, a critical consideration for aligning model outputs with clinical perspectives. While interpretability and fairness are essential for building trust in healthcare AI \cite{XAI_Health,Fair_ML}, existing studies that address these in UTI prediction predominantly utilize data from hospital or emergency department settings, limiting their generalizability to broader patient populations \cite{JAKOBSEN2023,XAI_UTI1}. Our work aims to address these open questions by utilizing a linked EHR dataset that includes data from primary care, secondary care, and pathology data, allowing for a more comprehensive analysis of UTI risk. This dataset encompasses data from approximately one million individuals in the Bristol, North Somerset, and South Gloucestershire regions of the UK. Using this extensive dataset, we aim to better characterize UTIs and identify associated risk factors across various subgroups. 

Considering the limited availability of labeled UTI data, we propose a UTI risk estimation framework informed by clinical expertise. This framework enables automatic annotation of patient records and UTI risk estimation across individual patient timelines. We examine UTI risk predictors over a one-year observation period, highlighting how these predictors differ across UTI risk groups and contribute to increased risk while ensuring interpretability and transparency.

We detail the unique dataset we have acquired and associated challenges in Section \ref{sec:2}. In Section \ref{sec:3}, we present data pre-processing and curation strategies that transform raw EHR data into structured representations suitable for further statistical analysis and AI modeling. Section \ref{sec:4} introduces our clinician-based risk and outcome estimation framework. The methods for constructing a baseline model to identify initial predictive factors across different UTI groups are presented in Section \ref{sec:5}, while Section \ref{sec:6} presents the results from machine learning models and explainable AI techniques. Finally, we discuss limitations of this study and outline future research directions in Section \ref{sec:7}.

\section{Dataset}
\label{sec:2}

The Bristol, North Somerset, and South Gloucestershire (BNSSG) 'SystemWide' dataset consists of routinely recorded healthcare data \cite{BNSSGgithub}, with patient-level linked records primarily covering the period from October 2019 to July 2022. This dataset includes data on 962,237 adult patients whose Primary Care practices opted into data sharing within the BNSSG Integrated Care Board (ICB) area. Included in the dataset are records from primary care, secondary care, and notably pathology lab results for antibiotic susceptibility tests, with our specific extraction focused on UTI risk factor analysis and prediction modelling. The dataset was linked at the ICB and disseminated to the University via patient pseudo-anonymized identifiers and a strict governance regime to ensure the protection of patient privacy.

The primary care data in the extraction include patient demographics, living circumstances, and comorbidity histories (e.g., diabetes, dementia), as well as records of prescriptions dispensed for antibiotics, steroids, hormone replacement therapy, and catheter supplies. Secondary care data include information on hospital admissions and discharges, categorized by ICD-10 \cite{ICD10} codes for bacterial infections, and OPCS codes \cite{OPCS} for procedures related to chemotherapy, and urinary and gastrointestinal health. Pathology data include laboratory test results, including viral identification, blood, and urine culture tests. Further details on the information included and the demographic characteristics of the population in the dataset can be found in sections A1 and A2 in the Appendix. 

Like most EHR data, this real-world dataset presents several challenges that affect statistical analysis and machine learning modelling:

\begin{itemize}
\item \textit{Heterogeneity}: The BNSSG dataset includes both structured data (e.g., demographic information) and semi-structured data (e.g., open-ended categories for medicine and catheter information), reflecting the diversity of real-world EHR data. This variety complicates further analysis, as these data require distinct pre-processing techniques, and integrating these data types can introduce inconsistencies that affect the robustness of model training and interpretability.
\item \textit{Temporal Misalignment}: Data points are recorded at differing intervals, leading to time gaps and inconsistencies, as some events are recorded only when they occur (e.g., primary care medicine dispensations) while others are recorded at regular intervals (e.g., monthly records on co-morbidity status). This misalignment challenges the construction of time-dependent models and accurate analysis. Moreover, different patients contract UTIs at different times, which can be influenced by extrinsic factors (e.g. seasonality).
\item \textit{Sparsity}: Since certain data points are recorded only when healthcare contact events occur, there are often large temporal gaps, limiting the detail available to build complete patient profiles. Sparse data complicates statistical analysis and reduces the reliability of machine learning models, particularly for time-series analysis or cases requiring a consistent record of patient health.
\item \textit{Missingness}: Missing values in EHR - such as incomplete prescription data or catheter details in the BNSSG dataset - create challenges. Common approaches like imputation may add noise or bias if the causal context is not considered precisely, affecting the reliability of both the analysis and model predictions. This is notably a key concern when the data is not missing at random, such as when clinical tests are only conducted when disease is suspected, and then only if there is a substantive possibility that the result might trigger a change of treatment.  
\item \textit{Lack of an Explicit UTI Diagnosis Label}: Although the extracted data focus on UTIs, the SNOMED code \cite{SNOMED} diagnoses from primary care GPs were not available due to data governance limitations in our data transfer. Nevertheless, much UTI diagnosis in primary care is empiric and none-definitive. This lack of labeling makes supervised learning difficult, requiring us to rely on indirect indicators or semi-supervised approaches, which may be less accurate and increase the need for expert validation.
\end{itemize}

To address these challenges, we have collaborated closely with clinical partners to enhance data quality and transform the raw data into a consistent and suitable structure for machine learning, as detailed in the following sections.

\section{Data Pre-processing}
\label{sec:3}
Our data pre-processing pipeline prepares the large-scale EHR dataset for effective analysis by systematically cleaning, standardizing, and transforming data relevant to UTI prediction. Key steps include targeted cleaning of different data to ensure consistency across critical variables. We then discretise the data into days, transforming it into a sparse 3D matrix of subject by variable by day to support comprehensive but computationally efficient temporal analysis.

\subsection{Data Cleaning}
\label{sec:3.1}
With the focus on UTIs, we implemented targeted data cleaning procedures on specific data subsets, including primary care dispensation records, secondary care hospital admissions, and pathology lab results for blood and urine bacterial cultures. Each data source has unique characteristics that required consultation with clinical experts to inform tailored cleaning approaches.

\runinhead{Dispensed Dispensation Data} Relevant attributes were extracted from clinical prescription dispensation notes, including drug name, dosage, type (antibiotics, steroids, hormones), and administration route (e.g., oral and topical). For catheter-related dispensations, data include manufacturer, catheter type, insertion method, size, and material and coating categories (e.g., silicone and latex).

\runinhead{Hospital Admission Data} Admission data were cleaned to extract associated ICD-10 and OPCS codes recorded at each patient visit, alongside entry and discharge dates. This ensured we captured all relevant diagnostic and procedural information per admission period.

\runinhead{Pathology Data} Blood and urinary bacterial culture results contained both culture tests (e.g. species detected) and a mix of antibiotic susceptibility tests (ASTs) and other pathology results (114 unique test types). We categorized urinary bacterial culture results into “no significant growth”, “no growth”, “mixed growth”, “invalid”, “other” (non-AST), and “refer to AST test”. The reason for adding the “refer to AST test” category is because the urinary bacteria test data have AST results but coded in a non-accessible format that is more easily accessed in the AST data. AST results were denoted to “susceptible”, “intermediate”, and “resistant” responses. We grouped intermediate results under the resistant category to streamline analysis based on clinician guidance. Additionally, in non-AST cases, results were labeled as “positive” or “negative”. Both urinary culture and AST data include specimen source information, which we mapped to standardized categories such as catheter stream urine, mid-stream, and some other categories.

\runinhead{Demographic Data} All patients have monthly recorded demographic information including age, gender, sex, date of death, living circumstances, a set of comorbidities, and geographic location, using 2011 Lower Layer Super Output Areas (LSOAs) \cite{lsoa}. 

\subsection{Data Transformation}
\label{sec:3.2}
\begin{figure}[ht]
\centering
\includegraphics[scale=.6]{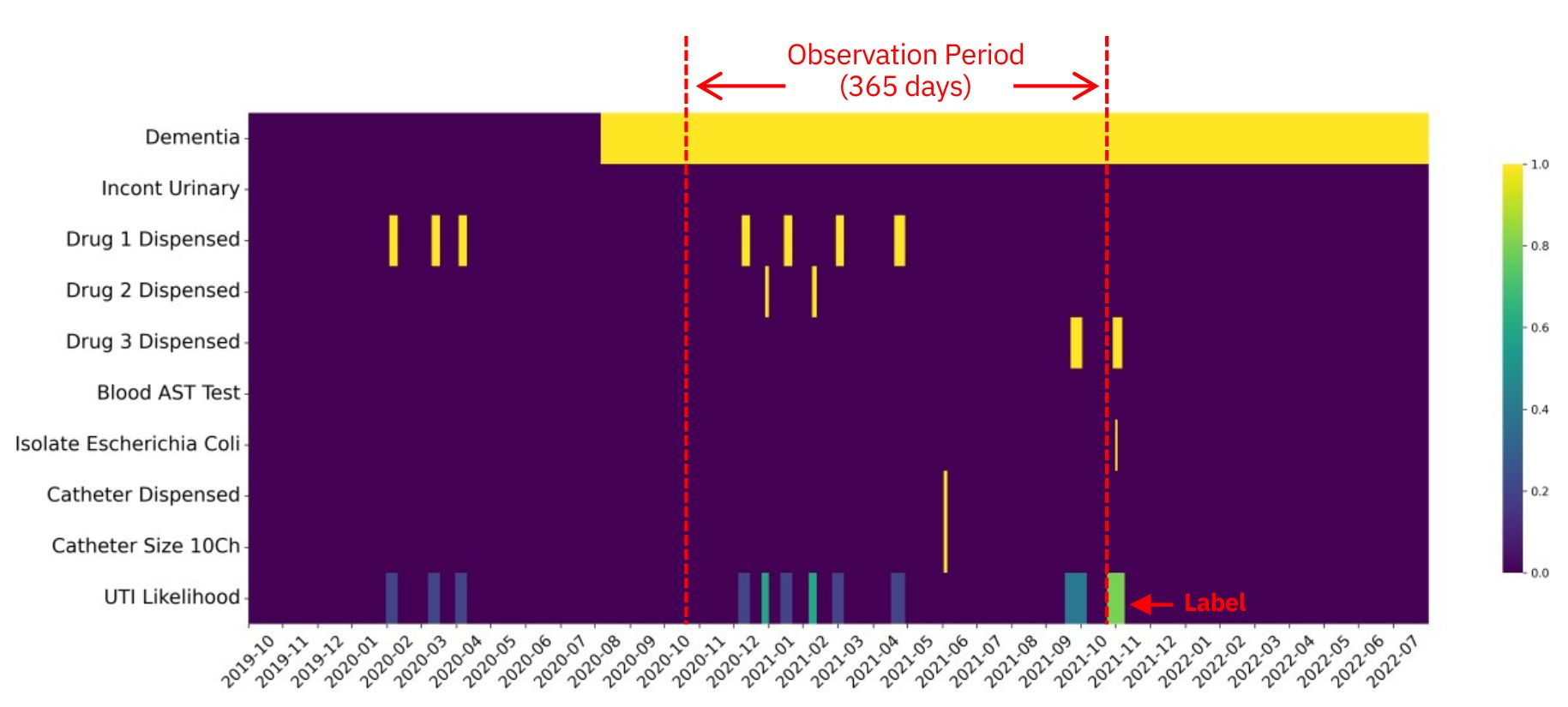}
\caption{A synthetic example of patient clinical events, with each feature represented as a binary value (0 or 1) indicating the presence (1) or absence (0) of an event. The outcome (UTI likelihood) consists of risk scores ranging from 0 to 1, with the latest highest risk score selected as the label. Features within the observation period are chosen for modeling.}
\label{ArtificialTimeline}       
\end{figure}

Following initial cleaning, we discretized the data into day-level resolution. The data are structured into a 3-dimensional matrix \(X(T,F,P)\), where \(T\) represents the time dimension, \(F\)  represents features, and \(P\)  denotes patients. This transformation is computationally intensive, as each patient’s timeline generates approximately 1,005 rows (one row per day over 33 months) and over 25,000 variables (e.g. a row for each unique ICD-10 code and every antibiotic and bacteria AST combination dedicating presence or absence). Figure \ref{ArtificialTimeline} provides an artificial example of a patient's clinical timeline after data processing, displayed as a heatmap. 

%

\section{UTI Risk Estimation Framework}
\label{sec:4}
Urinary tract infections can be challenging to diagnose due to the variety of symptoms that overlap with other conditions and the lack of a definitive diagnostic test. Given the absence of a direct UTI label in our dataset, we developed a UTI risk estimation framework in collaboration with clinical experts. Author PW is a NHS consultant in infectious disease and AH is primary care physician and professor of medicine. Both have extensive clinical experience dealing with UTI patient diagnosis and treatment. Drawing on this experience they were able to advise the data scientists on how construct the framework table below.

\begin{figure}[t]
\centering

\includegraphics[scale=.5]{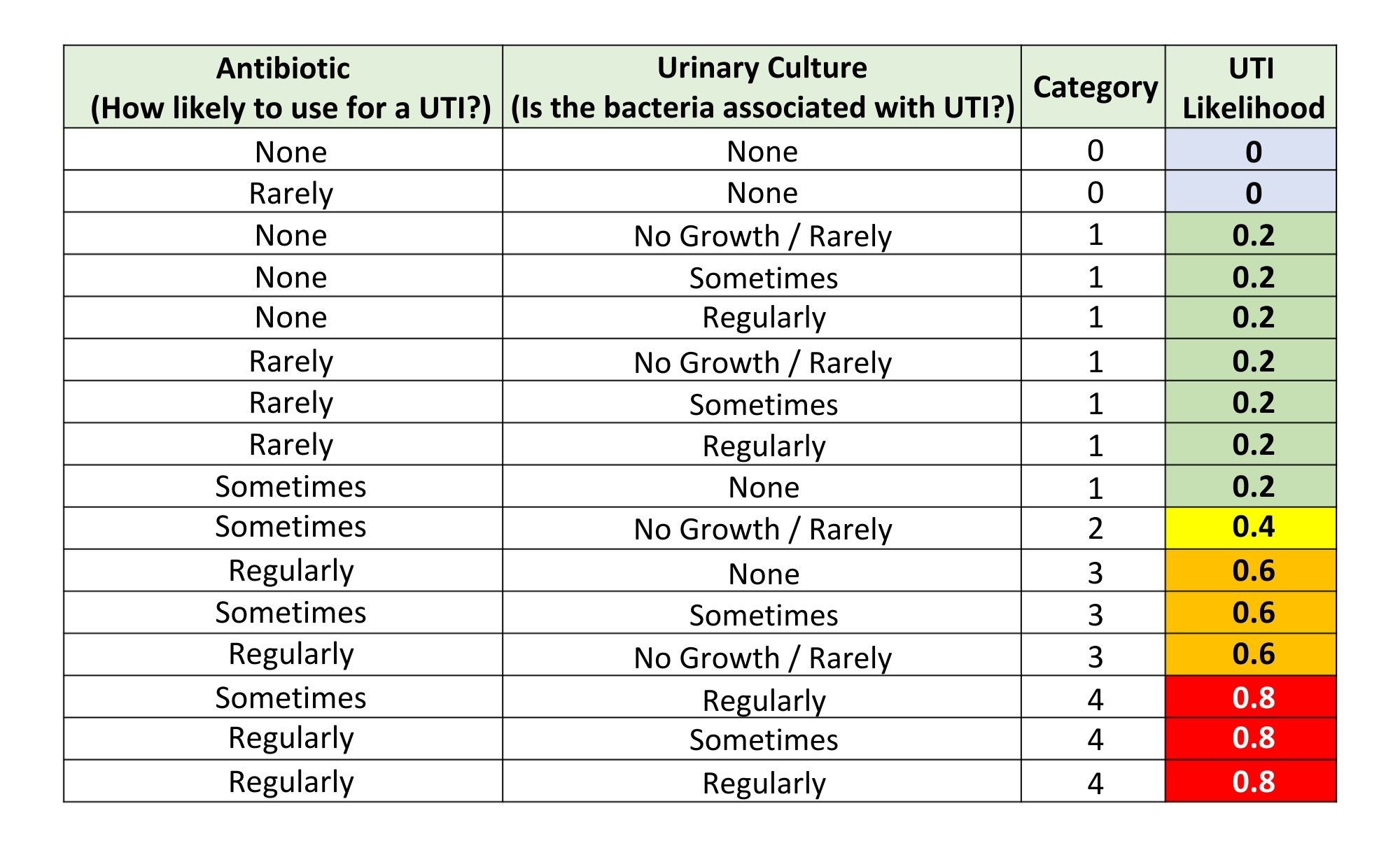}

\caption{UTI risk estimation framework.}
\label{framework}       
\end{figure}

This framework aims to provide a probabilistic estimate of UTI risk for individual patients, ranging from 0 (no UTI) to 1 (definite UTI). It is based on three key variables: (1) the presence of UTI-related antibiotic dispensations, (2) the identification of UTI-related bacteria in urine culture tests, and (3) relevant hospital diagnosis codes. By integrating these sources of information, our framework allows us to assess UTI risk across a patient’s entire clinical timeline, providing day-specific risk estimates. The details of the framework can be found in Figure \ref{framework}. 

\runinhead{Antibiotic Dispensation} Antibiotics dispensed to a patient are categorized into four groups based on their relevance to UTI treatment: "\textit{None}", "\textit{Rarely} Used", "\textit{Sometimes} Used", and "\textit{Regularly} Used" for UTI. The classification reflects the inferred levels that the prescribed antibiotic was used to treat or prevent a UTI.

\runinhead{Urine Pathology Results} Urine culture results are categorized into four levels: "\textit{None}", "\textit{No Growth} / Bacteria \textit{Rarely} Associated with UTI", "Bacteria \textit{Sometimes} Associated with UTI", and "Bacteria \textit{Regularly} Associated with UTI". We treat both "No Growth" and "Bacteria Rarely Associated with UTI" as equivalent, since having a urine specimen taken indicates that the patient likely had some UTI symptoms, but both indicate a low (but non-zero) likelihood of UTI presence.

\runinhead{Hospital Diagnosis Codes} Secondary care ICD-10 diagnosis code N39.0, which specifically corresponds to UTI, is a definitive source of UTI diagnosis. Although patients who are hospitalized are assigned this ICD-10 code, the occurrence of such cases is relatively rare in our dataset, with approximately 1,000 instances recorded during the study period. This is likely due to many UTI cases being managed in primary care or outpatient settings, rather than requiring hospitalization. 

These three variables are combined to generate a composite "UTI Likelihood", which provides a discrete likelihood score with values of 0, 0.2, 0.4, 0.6, 0.8, or 1, representing varying degrees of UTI risk, from none to confirmed.

\begin{figure}[t]
\sidecaption[t]
\includegraphics[scale=.33]{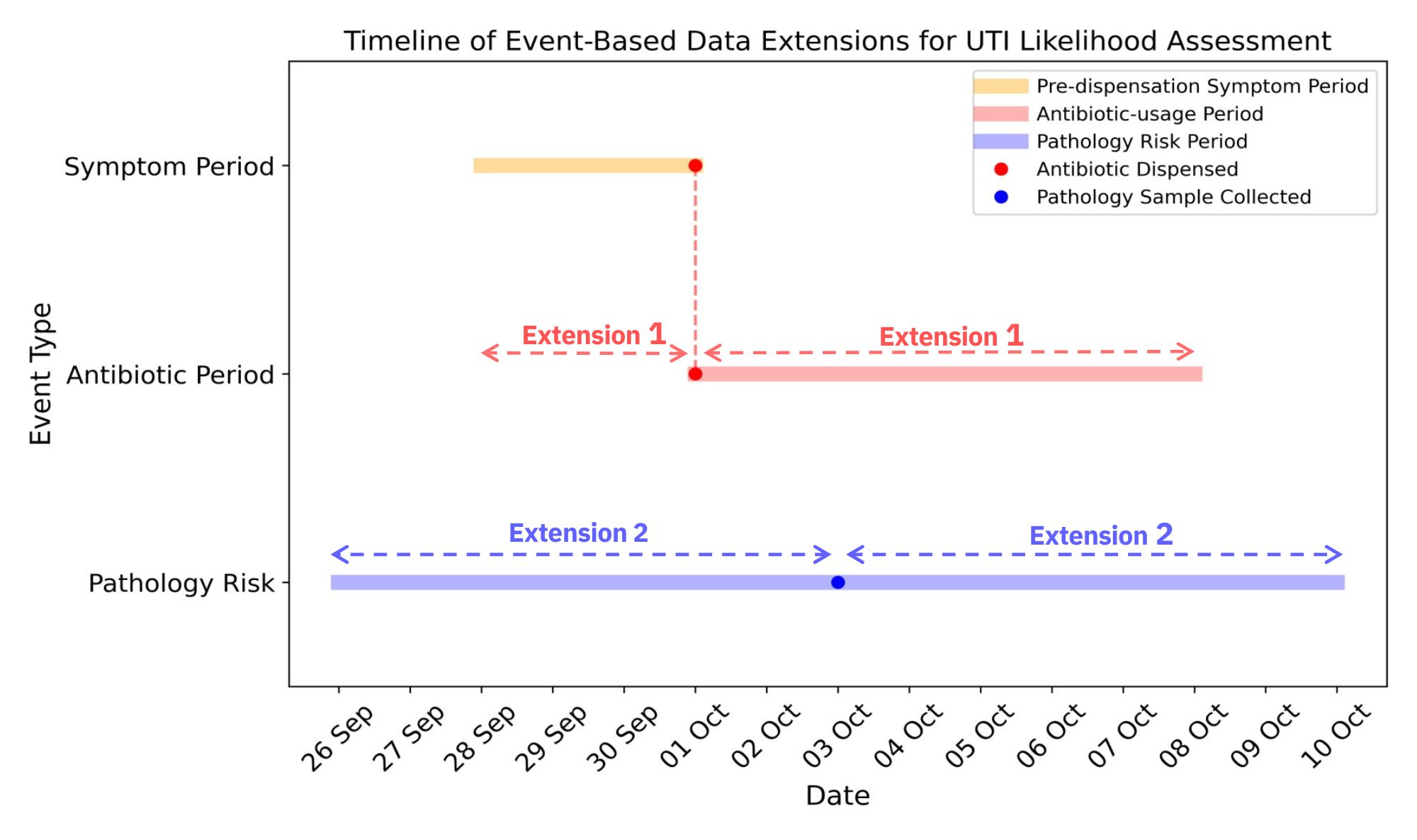}
\caption{Example of data extensions showing a patient on a 7-day antibiotic regimen (treatment duration may differ in other cases, based on individual prescription).}
\label{extension_plot}       
\end{figure}

One of the primary challenges in applying this framework is the misalignment of EHR data. The key variables - antibiotic dispensation and pathology results - are event-based records, meaning they are recorded at specific points in time (e.g., the day an antibiotic was dispensed, or a urine sample was processed). These events are not always aligned with each other, even when they may be causally related to an underlying UTI. To address this, we extend the data temporally to capture the ongoing effects of medications and the progression of infection, thereby enabling the application of our clinician-based risk framework. Figure \ref{extension_plot} illustrates an example of data extensions to achieve temporal alignment of events.

\runinhead{Data Extension 1} 
Given that a patient may have been symptomatic for days before receiving an antibiotic dispensation, we extend the data by marking a 3-day window prior to dispensation with the same antibiotic category based on clinical expertise. This accounts for the possibility that patients were already experiencing UTI symptoms before visiting a healthcare provider. Additionally, we estimate the duration of medication-usage based on prescribed dosage guidelines. For instance, if a patient is dispensed 28 tablets of Trimethoprim (500mg), we can infer the likely duration of treatment using local clinician guidelines or the World Health Organization’s defined daily dosage (DDD) guidelines for antibiotics \cite{DDD}. This enables us to extend the presence of the antibiotic over a period of days, reflecting the ongoing impact of the medication on future UTI risk. 

\runinhead{Data Extension 2}
For pathology records indicating that a urine specimen was collected, we extend the potential UTI risk to cover a 7-day window before and after the specimen collection date according to clinical advice. This extension accounts for the time required for pathogens to grow, be detected, and treated in the culture.

Secondary care records, including hospital admissions and discharges, are marked with specific entry and exit dates. These dates are used to denote the period during which a UTI diagnosis occurred. However, since these records do not provide detailed temporal alignment for antibiotic treatment or pathology results, we do not extend these records in time.

The extended data generated through these strategies allows to apply the framework more effectively, capturing a full picture of each patient’s clinical journey for a more robust risk assessment. Examples using the framework with data extensions applied to infer the UTI likelihood are illustrated in Appendix A3.

\section{Methodology}
\label{sec:5}
\subsection{Data Preparation}
During the 33-month data collection period (from October 1, 2019, to July 1, 2022), multiple potential occurrences of UTIs are identified for some patients, each assigned a likelihood score using the clinician-informed UTI risk estimation framework. As higher likelihood scores reflect greater confidence in the label, to boost the relatively small class sizes for the higher likelihoods, we selected the UTI occurrence with the highest likelihood within each patient’s observation window, favoring the most recent and confident estimate. For each patient, we extracted features from the 12-month period preceding this highest likelihood UTI event, as shown in Figure \ref{ArtificialTimeline}. This standardized observation window across all patients, enabling consistent identification of predictive risk factors influencing the likelihood of UTI occurrence within the next year. Features included in this study are listed in Appendix A5.

For the control group, consisting of individuals with a UTI likelihood of 0, we sampled individuals in a manner that aligned with the temporal distribution of the UTI group (likelihood \(>\) 0) across the years. This ensures that both the UTI and control groups had similar distribution of observation time-frames, facilitating a fair comparison between the two groups with respect to the distribution of timelines across the years and seasons.

To maintain uniformity in temporal analysis, patients with observation windows shorter than 12 months were excluded. Additionally, patients under the age of 18 and those with unknown or unspecified sex were removed to establish a more homogeneous cohort for analysis. The study workflow is described in Appendix A4.

\subsection{Modeling UTI Risk Categories}
The clinician-informed UTI risk estimation framework categorizes patients into six distinct likelihood groups. To analyze predictive differences between adjacent risk groups and identify key features predicting transitions between categories, we developed pairwise classification models using XGBoost \cite{xgboost} and SHAP (SHapley Additive exPlanations) \cite{SHAP} for enhancing explainability. XGBoost was selected based on its superior predictive performance compared to logistic regression and random forest models in prior experiments. SHAP was chosen for its strong theoretical foundation based on Shapley values and its robustness and consistency across models, ensuring reliable insights into feature contributions and interactions. Each pairwise models were designed to differentiate neighboring risk groups (e.g., 0 vs. 0.2, 0.2 vs. 0.4), with the higher-risk group within each pair considered the positive class. Data for each model was divided into training, validation, and testing sets with the ratio of (3:1:1) with the modeling workflow and hyperparameter details provided in Appendix A4. As the 0.4 and 1 UTI groups are relatively small, there is potential of class imbalance when modeling adjacent classes. To mitigate this issue, we experimented with three approaches: (1) no intervention, (2) applying the scale\_pos\_weight hyperparameter in XGBoost, and (3) employing SMOTE (Synthetic Minority Over-sampling Technique) \cite{SMOTE} for oversampling the minority class. While all three methods showed consistent Shapley feature explanations, the weighted balancing (using scale\_pos\_weight) was ultimately chosen as in our study. This method offered a good balance between model performance and stability in feature importance, without introducing the alterations in feature distributions observed with SMOTE. The pairwise models identify the important features distinguishing patients between adjacent UTI risk categories, providing insights about factors associated with elevated UTI risk. By leveraging SHAP explainable models to assess the impact of individual features on model predictions, we aim to align machine learning outputs with clinical understanding to ensure interpretability and transparency.

\section{Results}
\label{sec:6}
\subsection{Descriptive Analysis}
The study population consisted of individuals categorized into UTI and control groups, with key demographic characteristics summarized in Table \ref{tab:demo-uti&control}. Patients in the UTI group tend to be older on average than those in the control group, and a higher proportion of females are observed in the UTI group, reflecting the known epidemiology of UTIs.  Additionally, the UTI group had a greater prevalence of comorbidities, such as dementia. In terms of living conditions, a higher proportion of patients in the UTI group were reported as housebound or residing in a nursing or caring home, while these proportions were much lower in the control group. 

\begin{table}
\caption{Demographic Information of UTI and Control Groups.}
\centering
{
\begin{tabular}{@{}lccccccc@{}}
\toprule
{\textbf{Demographics}} & \multicolumn{7}{c}{\textbf{Groups}} \\ \cmidrule(lr){2-8}
 & \multicolumn{1}{c}  {\textbf{Control Group\footnotemark[1]}} & \multicolumn{6}{c}{\textbf{UTI Group\footnotemark[2]}} \\ \cmidrule(lr){2-2} \cmidrule(lr){3-8}
 & \textbf{Overall} & \textbf{Overall} & \textbf{0.2} & \textbf{0.4} & \textbf{0.6} & \textbf{0.8} & \textbf{1.0} \\ 
 \textnormal{ \hfill N = } &  \scriptsize \text{610602} & \scriptsize \text{147518} & \scriptsize \text{ 84280 } &\scriptsize \text{ 1969 } & 
 \scriptsize \text{ 38278 } & \scriptsize \text{ 21539 } & \scriptsize \text{ 1452 }\\ \midrule

\textbf{Age (years) (\%)} & & & & & & & \\ 
18-24 & 12.0 & 6.4 & 5.8 & 5.9 & 9.0 & 4.5 & 1.2 \\
25-44 & 42.4 & 33.8 & 38.9 & 37.0 & 31.2 & 19.6 & 5.9 \\
45-64 & 29.3 & 25.5 & 25.8 & 26.5 & 26.4 & 23.6 & 14.2 \\
65-84 & 14.3 & 26.5 & 23.5 & 24.3 & 25.0 & 39.9 & 45.0 \\
85+ & 1.9 & 7.7 & 5.9 & 6.3 & 8.3 & 12.4 & 33.6 \\ \midrule
\textbf{Gender (\%)} & & & & & & & \\ 
Male & 55.8 & 30.6 & 37.8 & 36.4 & 19.1 & 20.6 & 51.4 \\
Female & 44.2 & 69.4 & 62.2 & 63.6 & 80.9 & 79.4 & 48.6 \\ \midrule
\textbf{Comorbidities (\%)} & & & & & & & \\ 
Incontinent Urinary & 0.1 & 0.4 & 0.2 & 0.3 & 0.5 & 0.7 & 0.9 \\
Dementia & 0.6 & 3.2 & 2.0 & 1.7 & 4.7 & 4.6 & 12.8 \\
Covid High Risk & 3.8 & 15.2 & 15.2 & 16.6 & 13.3 & 16.9 & 38.8 \\
Covid Increased Risk & 30.3 & 57.4 & 55.0 & 56.6 & 54.8 & 69.2 & 91.9 \\
Organ Transplant  & \(<\)0.1 & 0.1 & 0.1 & 0.4 & 0.1 & 0.1 & 0.3 \\ \midrule
\textbf{Living Conditions (\%)} & & & & & & & \\
Housebound & 0.7 & 4.6 & 3.4 & 5.1 & 5.3 & 6.8 & 22.0\\
Nursing/Caring Home & 0.4 & 2.2 & 1.3 & 1.9 & 3.3 & 3.7 & 7.1 \\
Homeless & 0.1 & 0.1 & 0.1 & 0.2 & 0.1 & \(<\)0.1 & 0.1 \\
\bottomrule
\end{tabular}
}
\label{tab:demo-uti&control}
\end{table}
\footnotetext[1]{Individuals with a UTI likelihood \(=\) 0 }
\footnotetext[2]{Individuals with a UTI likelihood \(>\) 0}

\begin{figure}[!ht]
\includegraphics[scale=.26]{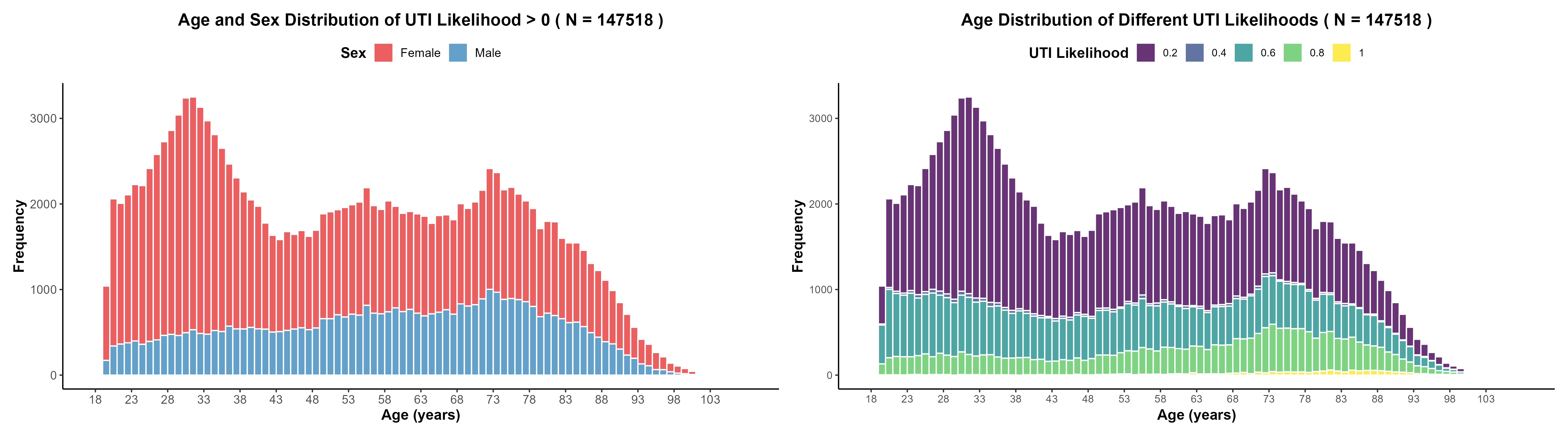}
\caption{Age and sex distribution of the UTI group (likelihood \(>\) 0).  (\textit{Left}) Age and sex distribution. (\textit{Right}) Age distribution of different UTI likelihood groups.}
\label{Uti_Dist}       
\end{figure}

\begin{figure}[t]
\includegraphics[scale=.29]{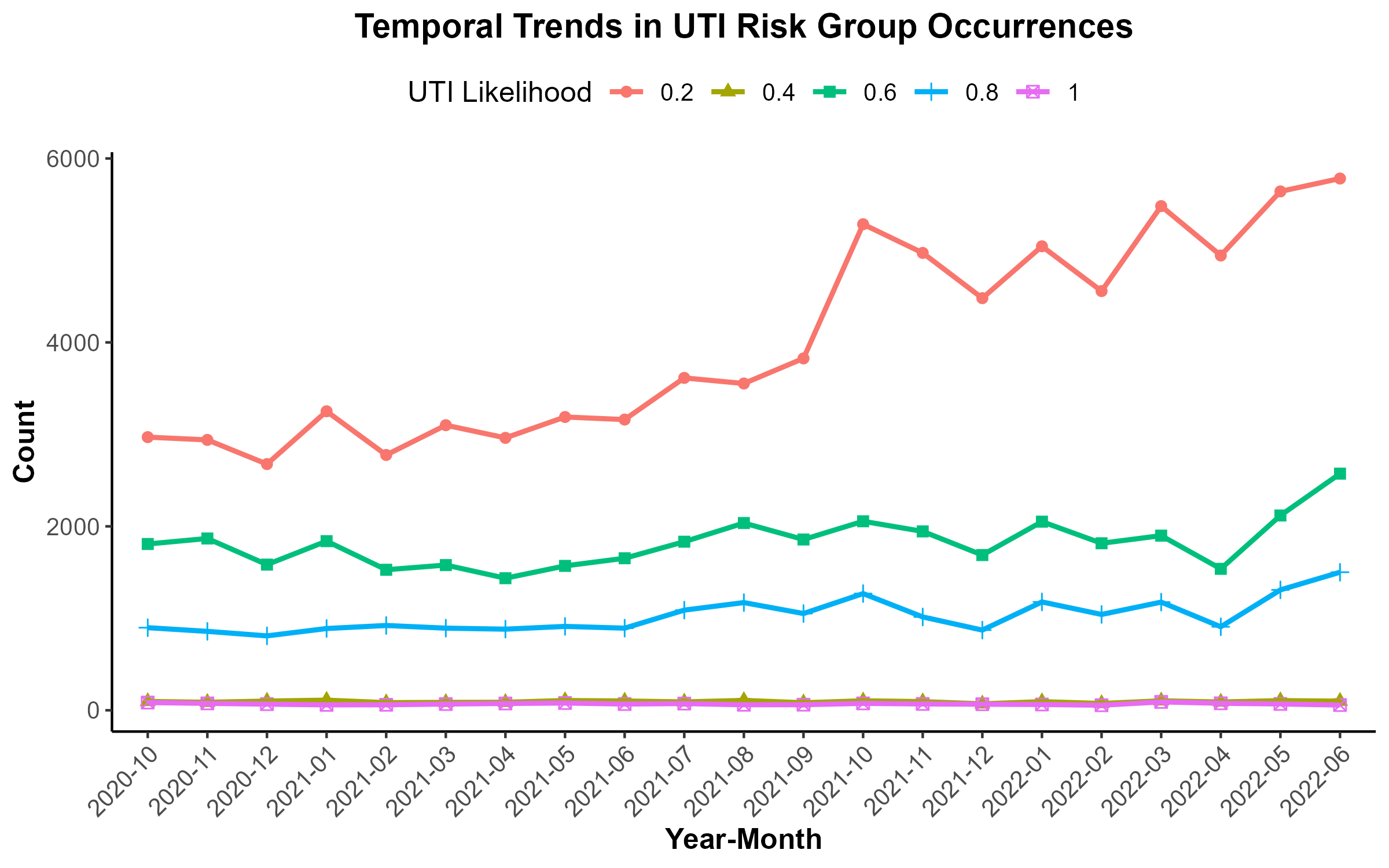}
\sidecaption[t]
\caption{Temporal trends in UTI risk group occurrences.}
\label{Uti_Trend}       
\end{figure}

Figure \ref{Uti_Dist} provides an overview of UTI distribution across age and gender, while the temporal distribution of patients across the five UTI likelihood categories (0.2 to 1.0) is shown in Figure \ref{Uti_Trend}. 

An analysis of age and gender distributions reveals a distinct peak among females around the age of 30 within the 0.2 risk group, a pattern not observed in males. This observation may reflect demographic influences on clinical practices, such as urine samples collected during pregnancy, where treatment is provided only if symptomatic or a pathogen is identified. The 0.2 risk group presents a significant proportion, while the 0.4 risk group is small, comprising patients who are dispensed antibiotics sometimes used for UTIs and with negative pathology results or findings involving bacteria rarely associated with UTIs. These risk groups capture cases where clinical suspicion of UTI exists, but uncertainty remains, resulting in a low likelihood risk score. The likelihood of 1.0 group, which corresponds to the highest certainty of UTI diagnosis (e.g., hospitalized patients who had ICD-10 codes confirming UTI), was limited by the relatively small number of severely ill patients in the dataset. From the temporal trend perspective, the 0.2 group shows an obvious increase after mid 2021. This may be attributed to challenges in accessing healthcare during the pandemic and a subsequent rise in clinical activities as restrictions were eased. Interestingly, a slight upward trend in the 0.6 and 0.8 groups after April 2022 is discovered, possibly reflecting the impact of policy changes or adjustments in clinical practices.

\subsection{Model Performance and Feature Importance}
\begin{figure}[t]
\includegraphics[width=0.45\columnwidth,height = 0.37\columnwidth]{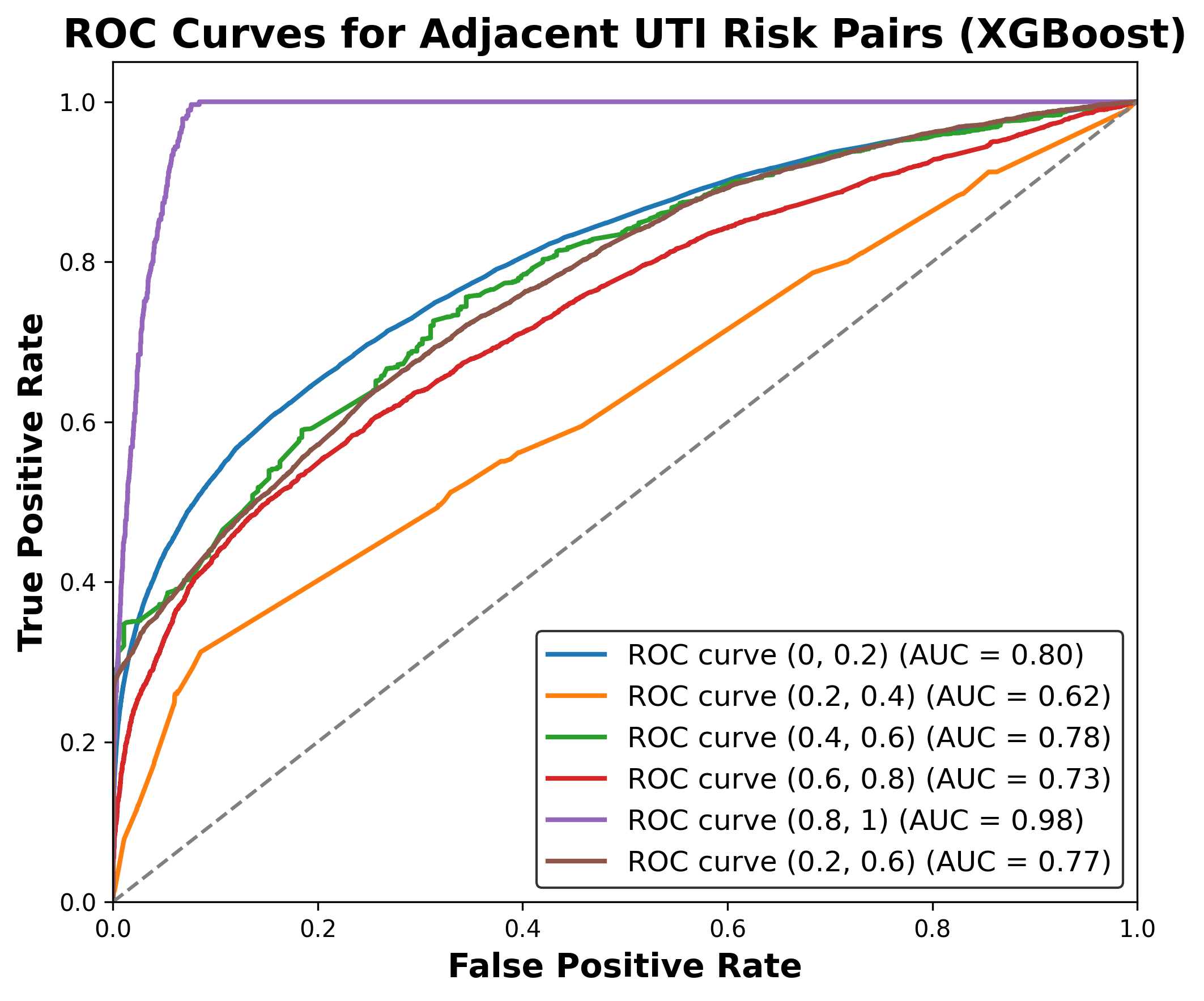}
\sidecaption[t]
\caption{Roc-curve for all the pairwise models over the testing data.}
\label{roc_curve}       
\end{figure}

To differentiate patients between neighbouring UTI likelihood categories, pairwise models were constructed for adjacent groups (e.g., 0 vs. 0.2, 0.2 vs. 0.4). However, from the clinical perspective underlying the UTI risk framework, the 0.4 risk group is notably small, with only one instance allocated to this category, making it challenging to model effectively. To address this, we constructed an additional pairwise model for the 0.2 vs. 0.6 groups to explore the differences between these two categories. Table \ref{tab:performance_metrics} summarizes the performance metrics of these models over the testing set, including accuracy, precision, recall, F1-score, and AUC-ROC. Figure \ref{roc_curve} displays the ROC curves for all the pairwise models, providing a visual comparison of their discriminative ability across the adjacent UTI risk categories. The models demonstrate varying effectiveness in distinguishing between adjacent UTI risk groups. The model differentiating higher-risk groups \textbf{0.8 vs. 1.0} generally performs better, with high accuracy (0.94) and AUC-ROC (0.98), suggesting that these groups are more easily distinguishable. However, models distinguishing between lower-risk groups, like \textbf{0.0 vs. 0.2} and \textbf{0.2 vs. 0.4}, tend to show lower precision and F1-scores, which may be influenced by class imbalance, particularly the small size of the 0.4 group. The \textbf{0.2 vs. 0.6} model, introduced to explore the differentiation between a low and moderately high-risk group, shows moderate accuracy (0.71) but a relatively low F1-score (0.58), reflecting the challenge of distinguishing between groups that are distinct but still share similar characteristics. 

\begin{table}[t]
\centering
\caption{Performance metrics of the pairwise models.}
\label{tab:performance_metrics}       
%
%
\begin{tabular}{p{2cm}p{1.5cm}p{1.5cm}p{1.5cm}p{1.5cm}p{1.5cm}}
\hline\noalign{\smallskip}
\textbf{Model} & Accuracy & Precision & Recall& F1-score & AUC-ROC  \\
\noalign{\smallskip}\svhline\noalign{\smallskip}
\textbf{0.0 vs 0.2} & 0.81 & 0.34 & 0.61 & 0.44 & 0.80 \\
\textbf{0.2 vs 0.4} & 0.87 & 0.06 & 0.33 & 0.10 & 0.62 \\
\textbf{0.4 vs 0.6} & 0.69 & 0.98 & 0.69 & 0.81 & 0.78 \\
\textbf{0.6 vs 0.8} & 0.72 & 0.62 & 0.53 & 0.57 & 0.73 \\
\textbf{0.8 vs 1.0} & 0.94 & 0.51 & 0.94 & 0.66 & 0.98 \\
\textbf{0.2 vs 0.6} & 0.71 & 0.53 & 0.64 & 0.58 & 0.77 \\
\noalign{\smallskip}\hline\noalign{\smallskip}
\end{tabular}
\end{table}

\begin{figure}[!ht]
\centering
\includegraphics[width=1.0\columnwidth,height = 0.83\columnwidth]{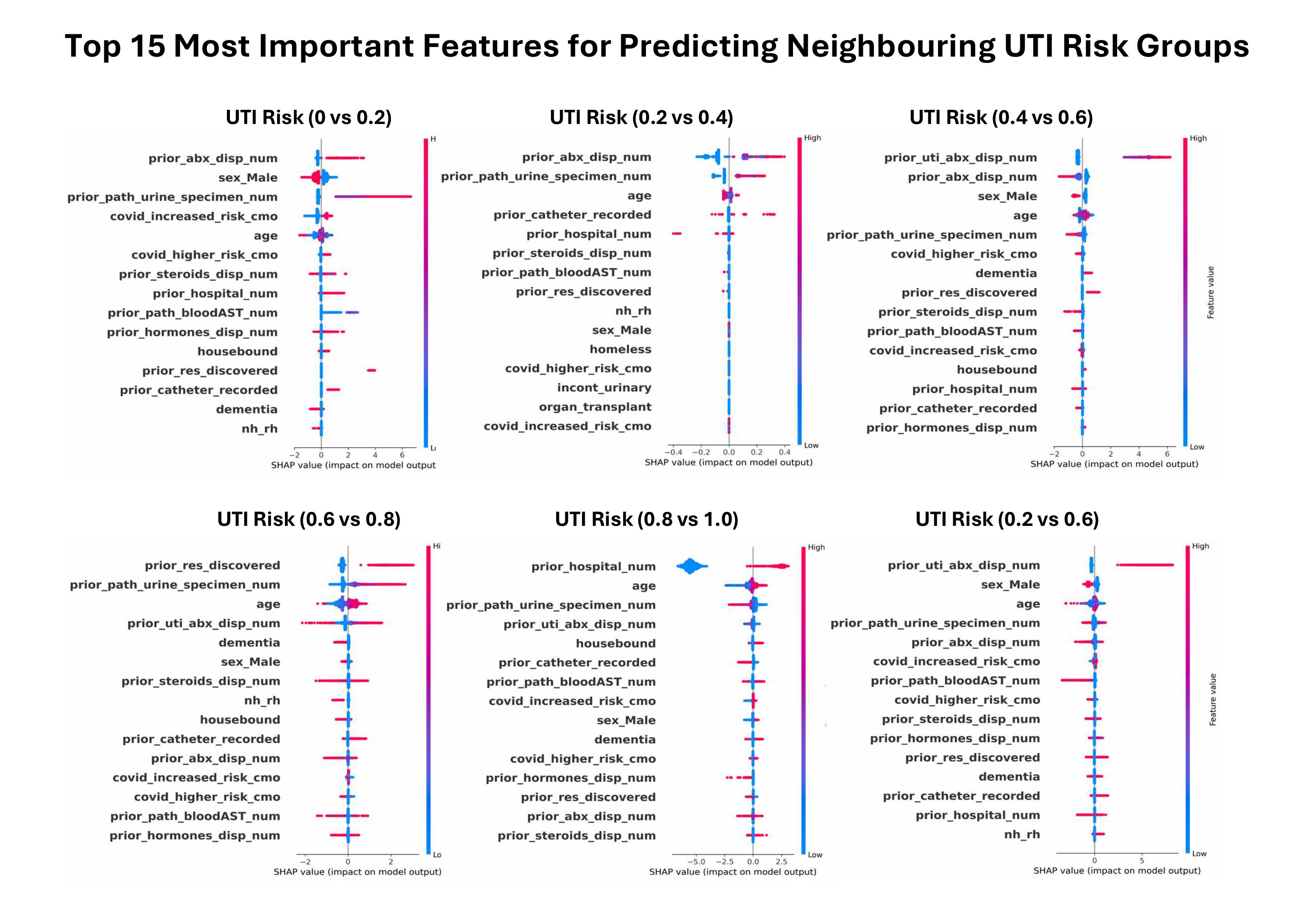}
\caption{Top 15 features identified by SHAP for the six pairwise XGBoost models distinguishing adjacent UTI risk categories.}
\label{SHAP}       
\end{figure}

The SHAP summary plots (Figure \ref{SHAP}) reveal evolving patterns in feature importance as the risk groups progress, reflecting both clinical and demographic factors that influence UTI likelihood. In the \textbf{0.0 vs. 0.2} model, features related to prior clinical interventions, such as the number of previous antibiotics dispensed and the number of urine specimens processed, play a prominent role in distinguishing between low-risk groups, suggesting that historical clinical events within one year contribute to predicting risk in these less certain cases. In the \textbf{0.4 vs. 0.6} model, the focus shifted towards more specific clinical features, including the number of UTI-specific antibiotics dispensed. Additionally, comorbidities like dementia and prior resistance discoveries become more relevant, indicating their growing importance in identifying patients at moderate risk. In the \textbf{0.2 vs. 0.6} model, which differentiates between low and moderately high-risk patients, the number of UTI-specific antibiotics dispensed again stands out as a key feature. For the \textbf{0.6 vs. 0.8} and \textbf{0.8 vs 1.0} models, age displays a clearer trend, with older patients more likely to be classified into the higher-risk groups. Features such as prior resistance discoveries and the number of urine specimens processed are particularly influential in the model of the \textbf{0.6 vs. 0.8}. When distinguishing patients in the group of \textbf{0.8 vs. 1.0}, the model highlights severe case predictors, such as housebound status and the number of previous hospital visits, underscoring the role of these factors in patients with hospital-confirmed UTI diagnoses (UTI likelihood \(=\) 1.0).

The results from SHAP help understand how the models prioritize different features at each stage of the risk estimation process. These observations suggest that the importance of various clinical and demographic features changes across risk categories, with certain factors becoming more prominent as risk increases. While these results provide useful insights, future work will be necessary to refine the categorization of risk groups, particularly by adjusting the groupings to better capture the variability in patient risk profiles and improve model robustness.

\section{Discussion}
\label{sec:7}

In this study, we introduced a unique linked EHR dataset and detailed the pre-processing and curation methods tailored for UTI prediction. Despite its potential to advance AI-driven healthcare solutions, the dataset has several limitations. The use of dispensation data, rather than prescribing records commonly used in similar studies, provides a more accurate representation of actual medication usage but lacks confirmation of whether patients initiated or completed their medication courses. Additionally, urine sample processing data may have temporal inconsistencies, as the processing date does not always align with the sample collection date. Our initial clinician-guided framework, while grounded in current clinical practices, focuses on a limited set of clinical indicators and may overlook subtle or unrecognized factors that influence UTI risk. Furthermore, the relatively small size of the 0.4 risk group may affect the robustness of analyses, suggesting the need to consider merging it with adjacent groups or exploring alternative groupings using AI-driven methods. Finally, addressing the challenges posed by incomplete data and the lack of explicit UTI labels remains critical. These limitations introduce uncertainty into the analysis, emphasizing the importance of robust estimation methods to enhance the reliability of risk assessment.

To address these limitations, future work will focus on refining the risk estimation framework through advanced sub-grouping and clustering techniques, guided by explainable AI, to uncover more comprehensive patterns in the data and enhance UTI risk group definitions. This includes exploring data-driven strategies for reclassifying smaller or less robust groups, such as the 0.4 risk category, to improve analytical reliability. Additionally, efforts will be directed toward investigating subgroup-specific risk factors to ensure fairness and inclusivity, particularly for underrepresented patient populations. Incorporating novel clinical indicators and integrating longitudinal patient histories may provide deeper insights into the predictors of UTI risk. Enhanced validation through external datasets and collaborations with diverse healthcare settings will further ensure the robustness and generalizability of the findings.

\section{Conclusion}
\label{sec:8}

This study represents a first step in developing a clinician-guided framework for UTI risk predictors investigation with explainable AI. While preliminary, our results offer insights into the predictors of UTI risk and demonstrate the importance of integrating data curation, risk estimation frameworks, and interpretable AI techniques. However, further validation and refinement are needed before these models can be implemented in clinical practice. By leveraging a diverse and enriched dataset, we aim to improve the reliability of predictive analytics, promote healthcare equity, and enhance clinical decision-making and patient outcomes.




\section*{Supplementary material}

# A  Appendix

## A.1 Overview of dataset

Table 1 describes the information included in the dataset and relevant data availability periods. Due to the insufficiency of community care data, we make minimal use of information from this source for our analysis.

*Table 1 Dataset overview and data availability periods*

| Data | Data Available Time |
|---|---|
| ***PRIMARY CARE*** | |
| **Demographics** | |
| Age | October 2019 to July 2022 |
| Gender | |
| Sex | |
| Date of Death (Month-Year) | |
| Living Circumstances (Homeless, Housebound, Care Home) | |
| Lower-Layer Super Output Area (LSOA) | |
| **Comorbidities** | |
| COVID-19 Risk (High or Increased) | March 2020 to July 2022 |
| Dementia | October 2019 to July 2022 |
| Incontinence Urinary | |
| Organ Transplant | |
| **Medicine Dispensation** | |
| Drug (Antibiotics, Steroids, and Hormone Replacement) Dispensation | October 2019 to July 2022 |
| Drug Dose | |
| Drug Quantity | |
| Drug Dispensation Description | |
| Catheter Dispensation | |
| ***SECONDARY CARE*** | |
| **ICD10 Diagnosis Code Related to:** | |
| Diagnoses for any bacterial infection | October 2019 to July 2022 |
| **OPCS Procedure Codes Related to:** | |
| Chemotherapy | October 2019 to July 2022 |
| Catheter Procedures | |
| Surgeries/Procedures Near Urinary Region | |
| ***PATHOLOGY*** | |
| Urine Bacterial Culture (Organism, Antimicrobial Susceptibility) | October 2016 to July 2022 |
| Blood Bacterial Culture (Organism, Antimicrobial Susceptibility) | |
| Virology (Organism) | |
| Clostridiodes Difficile Tests (GBH, Toxin, PCR) | |
| ***COMMUNITY CARE, visits related to catheters:*** | |
| Code / Reason for visiting | October 2018 to January 2020 |
| Person who changes the catheter | |

## A.2 Demographic information of the population

Table 2 shows the demographic information of the population in the dataset. Please note that the age displayed here is the first record in the demographic data within the data available timeline.

*Table 2 Demographic information of the whole population in the dataset*

| Category | Group | Frequency (%) |
|---|---|---|
| Age | 18--24 years | 153941 (16.0) |
|  | 25--44 years | 369806 (38.4) |
|  | 45--64 years | 257167 (26.7) |
|  | 65--84 years | 152702 (15.9) |
|  | 85+ years | 28621 (3.0) |
| Gender | Male | 484592 (50.4) |
|  | Female | 477613 (49.6) |
|  | Not Specified/Unknown | 32 (0.0) |
| Comorbidities | Incontinent Urinary | 1458 (0.2) |
|  | Dementia | 13130 (1.4) |
|  | Covid High Risk | 60962 (6.3) |
|  | Covid Increased Risk | 344394 (35.8) |
|  | Organ Transplant | 514 (0.1) |
| Living Conditions | Housebound | 16782 (1.7) |
|  | Nursing/Care Home | 10029 (1.0) |
|  | Homeless | 858 (0.1) |

## A.3 Examples of UTI Risk Likelihood Inference

This section presents several examples of UTI likelihood (L) inferred from evidence data using the risk estimation framework, along with two data extensions. Figure 1 depicts a hospitalized patient with an ICD-10 code of N39.0 from days 5 to 11, indicating a confirmed UTI diagnosis (L = 1.0). Figure 2 shows a patient who was dispensed Nitrofurantoin on day 5, with a urine culture result on day 10 revealing 'No Growth.' In this case, the UTI likelihood for the overlapping days of these two events is 0.6, as determined by the risk estimation framework. Figure 3 illustrates a patient dispensed Nitrofurantoin on day 5, with a urine culture result on day 10 showing the presence of *E. coli*. For this case, the UTI risk estimation increases to 0.8 for the overlapping days, as the identification of *E. coli*, a bacterium commonly associated with UTIs, raises the likelihood compared to the example shown in Figure 2.

*Figure 1 Example of a hospitalized patient*

*Figure 2 Example of a patient with Nitrofurantoin dispensed and 'No Growth' urine culture test result*

*Figure 3 Example of a patient with Nitrofurantoin dispensed and a urine culture test result of E. coli*

## A.4 Modelling Workflow and Hyperparameters

Table 3 describes the model hyperparameters used in each model. All other hyperparameters are set to their default values. Table 4 shows the overall study workflow.

*Table 3 XGBoost hyperparameters specific to each model.*

| Pairwise Models | Learning_rate | Max_depth | N_estimators | Scale_pos_weight |
|---|---|---|---|---|
| 0.0 vs 0.2 | 0.2 | 3 | 150 | 7.218750616018451 |
| 0.2 vs 0.4 | 0.01 | 3 | 50 | 41.767768595041325 |
| 0.4 vs 0.6 | 0.2 | 3 | 50 | 0.05220043572984749 |
| 0.6 vs 0.8 | 0.1 | 5 | 100 | 1.7603445623750191 |
| 0.8 vs 1.0 | 0.1 | 7 | 50 | 14.710706150341686 |
| 0.2 vs 0.6 | 0.1 | 7 | 100 | 2.192827059181104 |

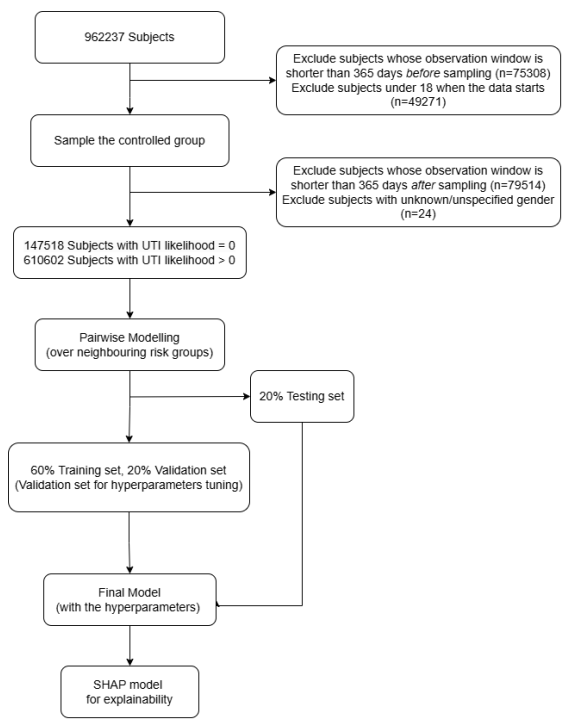

*Figure 4 Study workflow*

## A.5 Features Included in Modelling

| Variables | Type | Description |
| --- | --- | --- |
| Age | Numerical | The age of the patient when the UTI occurred |
| Sex | Category | The sex of the patient |
| Prior_antibiotics_num | Numerical | The number of any antibiotics dispensed to the patient |
| Prior_steroids_num | Numerical | The number of steroids medications dispensed |
| Prior_hormones _num | Numerical | The number of hormones medications dispensed |
| Prior_UTI_ABX_num | Numerical | The number of antibiotics specifically for UTI dispensed |
| Prior_resistance_discovered | Boolean | Whether the patient tested positive for resistance to any antibiotic |
| Prior_catheter_recorded | Boolean | Whether the patient has ever used a catheter |
| Prior_hospital_visits_num | Numerical | The number of hospital admissions the patient had |
| Prior_path_urine_specimen_num | Numerical | The number of urine specimens collected for pathology testing |
| Prior_path_bloodAST_num | Numerical | The number of blood specimens collected for aspartate aminotransferase (AST) tests |
| Incont_urinary | Boolean | Whether the patient has a recorded history of urinary incontinence. |
| Covid_increased_risk_cmo | Boolean | Whether the patient had risks of contracting Covid-19 prior to the UTI |
| Covid_higher_risk_cmo | Boolean | Whether the patient was identified as being at high risk for severe outcomes from Covid-19 |
| Organ_transplant | Boolean | Whether the patient is an organ transplant recipient |
| Dementia | Boolean | Whether the patient has been diagnosed with dementia. |
| Housebound | Boolean | Whether the patient has ever been housebound. |
| Nh_rh | Boolean | Whether the patient has ever lived in a nursing or care home. |
| Homeless | Boolean | Whether the patient has ever experienced homelessness. |


%
%
%

\end{document}